\pdfoutput=1

\documentclass[11pt]{article}

\usepackage{acl}

\usepackage{times}
\usepackage{latexsym}
\usepackage{bm} 
\usepackage{graphicx}
\usepackage{amssymb}
\usepackage{amstext}
\usepackage{multirow,multicol}
\usepackage{threeparttable}
\usepackage{makecell}
\usepackage{booktabs}
\usepackage{bbding}
\usepackage{arydshln} 
\usepackage{stfloats}
\usepackage{algorithmicx,algorithm}
\usepackage[noend]{algpseudocode}

\usepackage{comment}
\usepackage[T1]{fontenc}

\usepackage[utf8]{inputenc}

\usepackage{microtype}

\usepackage{inconsolata}

\title{SimulS2S-LLM: Unlocking Simultaneous Inference of Speech LLMs for Speech-to-Speech Translation}

\author{Keqi Deng\textsuperscript{1}, Wenxi Chen\textsuperscript{2}, Xie Chen\textsuperscript{2}, Philip C. Woodland\textsuperscript{1} \\
         \textsuperscript{1}Department of Engineering, University of Cambridge, Trumpington St., Cambridge, UK. \\
         \textsuperscript{2}Shanghai Jiao Tong University, Shanghai, China\\
         \texttt{\{kd502, pw117\}@cam.ac.uk} \\}

\begin{document}
\maketitle
\begin{abstract}
Simultaneous speech translation (SST) outputs translations in parallel with streaming speech input, balancing translation quality and latency.
While large language models (LLMs) have been extended to handle the speech modality, streaming remains challenging as speech is prepended as a prompt for the entire generation process.
To unlock LLM streaming capability,
this paper proposes SimulS2S-LLM, which trains speech LLMs offline and employs a test-time policy to guide simultaneous inference. SimulS2S-LLM alleviates the mismatch between training and inference by extracting boundary-aware speech prompts that allows it to be better matched with text input data. 
SimulS2S-LLM achieves simultaneous speech-to-speech translation (Simul-S2ST) by predicting discrete output speech tokens and then synthesising output speech using a pre-trained vocoder. 
An incremental beam search is designed to expand the search space of speech token prediction without increasing latency.
Experiments on the CVSS speech data show that SimulS2S-LLM offers a better translation quality-latency trade-off than existing methods that use the same training data, such as improving ASR-BLEU scores by 3 points at similar latency.
\end{abstract}

\section{Introduction}
Simultaneous speech translation converts input speech into translation output before the speech input utterance ends, enabling low-latency interaction \cite{zhang-etal-2024-streamspeech}. The translation can be text or speech, classified as Simul-S2TT or Simul-S2ST. Conventional Simul-S2ST use a cascaded approach that includes automatic speech recognition (ASR), machine translation (MT), and text-to-speech (TTS) \cite{1597243}. However, cascaded methods suffer from error
propagation and hinder joint optimisation \cite{deng2024lst}. 
In Simul-S2ST the model must decide when to emit translation tokens from incomplete speech input, which is challenging due to the continuous nature and the uncertain duration of spoken data\footnote{Simultaneous inference/translation means both speech input and output are streamed, in contrast to  work that only streams speech generation based on complete input speech.}. Recent work has begun exploring end-to-end (E2E) Simul-S2ST \cite{ma2024learning, zhang-etal-2024-streamspeech, barrault2023seamless}. However, leveraging large language models (LLMs), known for their remarkable performance across a wide range of tasks, in Simul-S2ST remains a challenge. 

Text-based LLMs have shown widespread success \cite{brown2020language, touvron2023llama2, dubey2024llama, ouyang2022training} and have been extended to handle speech \cite{chu2023qwen, tang2024salmonn, deng2024wav2prompt}, by prepending the speech as a prompt for LLM output generation and conditioning the LLM on the speech prompts.
However, this decoder-only architecture struggles with streaming, since all of the speech prompt is prepended beforehand, and all subsequent generated output attends to the speech prompts \cite{chen2024bestow}.
Therefore, online modifications must rely on previously obtained speech-text alignments to limit the speech accessible for each text token \cite{seide2024speech, tsunoo2024decoder}.


To address these challenges and enable the use of speech LLMs for simultaneous speech translation, this paper proposes SimulS2S-LLM, which to the best of our knowledge is the first work to apply LLMs for Simul-S2ST.
Moreover, SimulS2S-LLM aims to avoid restricting speech LLMs to specific streaming tasks, achieved via offline training.
SimulS2S-LLM adopts a test-time Wait-k strategy \cite{Ma2018STACLST} during inference to achieve simultaneous translation, allowing it to use only limited speech input as prompts to generate predicted translations. To alleviate the training-testing mismatch caused by offline training, SimulS2S-LLM leverages a continuous integrate and fire (CIF) mechanism \cite{9054250}  to extract a token boundary-aware speech prompt from the streaming encoder input. 
For Simul-S2ST, SimulS2S-LLM predicts target-language discrete speech tokens based on the LLM hidden states
and then synthesising output speech in the target language using a pre-trained vocoder.
An incremental beam search is introduced to expand the search space while avoiding additional latency.
The system is trained in an end-to-end fashion with a fixed text LLM.
The proposed SimulS2S-LLM was evaluated on a Common Voice-based Speech-to-Speech (CVSS) translation corpus, showing improved quality-latency trade-offs compared to existing Simul-S2ST methods, despite being trained offline. 

The main contributions of the paper are listed below:
\begin{itemize}
    \item SimulS2S-LLM, to our knowledge, is the first work extending LLMs to Simul-S2ST.
    \item With boundary-aware speech prompts, a novel offline training method is proposed, unlocking the Simul-S2ST capabilities of speech LLMs without restricting them to certain streaming tasks, aligning with the expectations of LLMs.
    \item Based on LLM multi-layer hidden states, incremental beam search is designed to expand the prediction search space of speech tokens.
    \item Extensive experiments were conducted, including comparisons of different methods to extract speech prompts.
\end{itemize}



\section{Related Work}

\subsection{Simultaneous Speech Translation}
\label{e2e-sst}
Existing simultaneous speech translation methods focus on speech-to-text translation (Simul-S2TT), which can be divided into fixed and flexible policies.
Wait-k is a typical fixed read-write policy that was initially proposed for text machine translation \cite{Ma2018STACLST} and then extended to speech translation \cite{Ma2020SimulMTTS, ren2020simulspeech, zeng-etal-2021-realtrans, dong-etal-2022-learning}. 
Furthermore, many studies have also explored flexible policy approach, including monotonic multi-head attention (MMA) \cite{Ma2020Monotonic}, the CIF-based method \cite{chang22f_interspeech}, neural transducers \cite{Xue2022LargeScaleSE}, and its variants \cite{deng2024lst, liu-etal-2021-cross, tang-etal-2023-hybrid}. These methods train the model in a streaming manner, enabling it to decide when to emit translation tokens on the fly.
Recently, some studies have explored using offline-trained attention-based encoder-decoder models for simultaneous inference \cite{liu20s_interspeech, papi-etal-2023-attention, papi23_interspeech}, such as determining whether to output translations based on attention scores \cite{papi-etal-2023-attention}. However, this strategy may pose challenges when applied to decoder-only architectures due to the reliance solely on self-attention.

\subsection{Direct Speech-to-Speech Translation}
\label{s2s}
Recent advancements in direct speech-to-speech translation have been driven by the use of discrete speech tokens, extracted from self-supervised pre-trained models such as HuBERT \cite{hsu2021hubert}.
The target-language discrete speech tokens are used as the training objective and vocoders are used to synthesise speech \cite{lee-etal-2022-direct}.
\citet{inaguma-etal-2023-unity} first transforms the source speech into hidden text states in the target language, based on which the target discrete speech tokens are generated.
\citet{DBLP:conf/iclr/DongH00KZF0WCYB24} uses cross-lingual LMs to convert source semantic tokens into target semantic tokens, which are then used to predict target acoustic tokens for speech generation.
Similarly, \citet{le2024transvip} jointly predicts the target-language text and residual vector quantisation codes.

Direct speech-to-speech translation is already very challenging, and performing it simultaneously (Simul-S2ST) requires
the translation to be generated based on incomplete source speech and is therefore a still harder task.
StreamSpeech \cite{zhang-etal-2024-streamspeech} uses connectionist temporal classification (CTC) \cite{graves2006connectionist} to align the source speech with the source text and target text, which are then used to guide simultaneous inference. It employs multi-task training, including ASR and Simul-S2TT tasks, to help in Simul-S2ST training. \cite{zhao2024textless} uses a neural transducer model \cite{Graves2012SequenceTW} to predict target-language discrete speech tokens from source speech. 
However, there is still a lack of research that  effectively leverages powerful LLMs for Simul-S2ST.

\subsection{Speech Large Language Models}
\label{speech-llm}
LLMs have achieved success \cite{achiam2023gpt, le2023bloom} and have been applied to text-based simultaneous translation \cite{DBLP:conf/emnlp/KoshkinS024,DBLP:conf/emnlp/KoshkinS024a}.
Several studies have extended LLMs to handle speech input \cite{chu2023qwen, zhang-etal-2023-speechgpt}. Speech LLMs\footnote{Further analysis of speech LLMs refers to \cite{10447605, deng2024wav2prompt}.} can be divided into two categories \cite{cui2024recent}. The first uses discrete speech tokens to extend the LLM vocabulary and build spoken generative LMs \cite{zhang-etal-2023-speechgpt, borsos2023audiolm, wang2023neural}. The second category uses continuous speech representations as the prompt to condition LLMs \cite{chu2023qwen, deng2024wav2prompt, 10447605, 10389705, 10445874, chen2023x, huang2024dynamic}.
This paper falls into the second category, as previous work \cite{fang2024llama} has shown that this approach can effectively leverage off-the-shelf text-based LLMs for efficient training.

With the advent of GPT-4o, speech-to-speech LLMs have attracted more attention and given rise to a series of models such as SpeechGPT \cite{zhang-etal-2023-speechgpt}. Mini-Omni \cite{xie2024mini} introduced parallel generation of text and audio, allowing models to initiate reasoning directly in audio. Llama-Omni \cite{fang2024llama} generates semantic speech tokens based on text hidden states.
Moshi \cite{defossez2024moshi} leverages both acoustic and semantic speech tokens to simultaneously model the input and output streams, enabling full-duplex operation. LSLM \cite{ma2024language} introduces a simplified way to achieve full-duplex operation using only semantic tokens. There are also some multi-modal generative LLMs, such as AnyGPT \cite{zhan2024anygpt}.
Our work differs by focusing on simultaneous inference, predicting target speech from incomplete source speech. In contrast,  prior work \cite{fang2024llama, xie2024mini, ma2024language} supports streaming speech generation but needs complete speech inputs or segments with sufficient information, whereas Simul-S2ST requires low latency and is thus more challenging.


\begin{figure}[t]
    \centering
    \includegraphics[width=77mm]{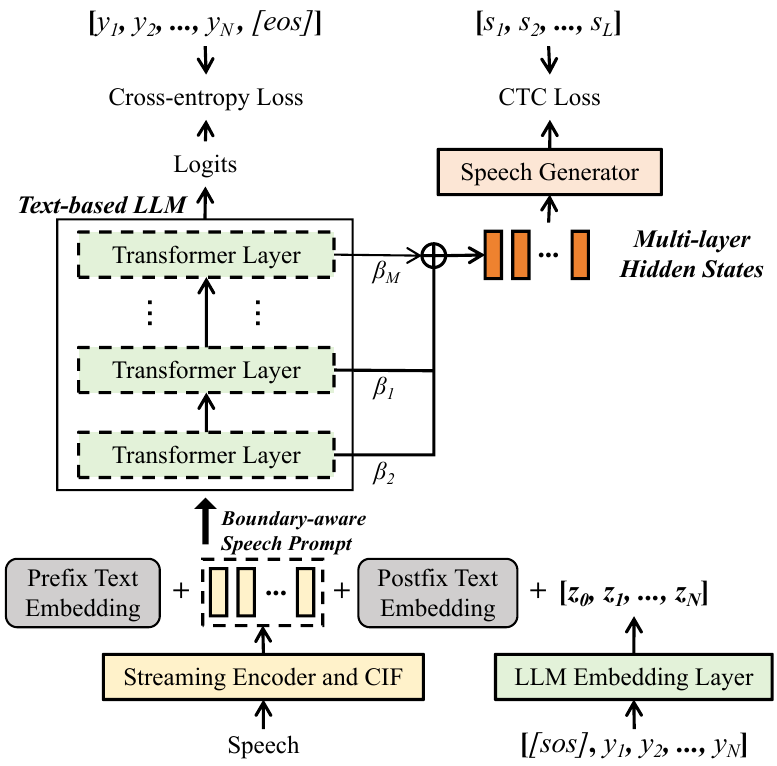}
    \caption{Illustration of SimulS2S-LLM offline training. $\bm{\oplus}$ denotes addition. Prefix and postfix text contain template instructions. The hidden states of each LLM layer are weighted ($\beta_i$) and summed.}
    \label{SimulS2S-LLM}
\end{figure}

\section{SimulS2S-LLM}
\label{method}

SimulS2S-LLM, as shown in Fig.~\ref{SimulS2S-LLM}, uses a streaming encoder to extract a boundary-aware speech prompt from the source speech and pre-pends it before the embeddings ($\bm{z}_0 \cdots \bm{z}_N$) of LLM text token input ($[sos], \cdots, y_N$), which follows a decoder-only architecture. The multi-layer hidden states of the LLM are weighted and summed, and the discrete output speech tokens ($s_{1} \cdots s_{L}$) are predicted in a streaming manner. SimulS2S-LLM is trained in an offline\footnote{Offline training refers to training where speech LLMs are not streaming-based, meaning that during training, the prediction of all tokens can attend to the entire speech input. This is independent of whether a streaming encoder is used.} manner, so that SimulS2S-LLM retains the potential to be applied to other non-streaming tasks.
After training, SimulS2S-LLM directly performs simultaneous inference for Simul-S2ST.

\subsection{SimulS2S-LLM Architecture}
SimulS2S-LLM contains four main modules: a streaming acoustic encoder, a CIF module, a text-based LLM, and a streaming speech generator. 
The encoder and the CIF are used to extract the boundary-aware speech prompts in a streaming manner, which is then fed into the text-based LLM along with other prompt templates that contain task instructions, i.e. the prefix and postfix text in Fig.~\ref{SimulS2S-LLM}, which are used to condition the text generation.

Training with teacher-forcing in LLMs can introduce a mismatch between training and testing, which may particularly affect the last-layer hidden state due to its focus on semantic information \cite{DBLP:conf/interspeech/ChangYFM023}. To address this, this paper employs a weighted sum of multi-layer LLM hidden states. Specifically, denote $\bm{h}^{m}_{i}$ as the hidden state of the $m$-th layer at the $i$-th step, the weighted sum is obtained with trainable weights $\beta_{m}$:
\begin{equation}
    \bm{h}_{i} = \beta_{1}\cdot\bm{h}_{i}^{1} + \cdots + \beta_{m}\cdot\bm{h}_{i}^{m}  \label{multi-layer}
\end{equation}

The hidden states $\bm{h}_{i}$ are fed into a streaming speech generator, which uses causal Transformer layers to enable streaming. SimulS2S-LLM employs semantic speech tokens as targets, with the predicted tokens passed to a vocoder to synthesise speech in the target language. Inspired by \cite{DBLP:conf/emnlp/SahariaCSN20, fang2024llama}, SimulS2S-LLM up-samples $\bm{h}_{i}$ and applies a simple CTC objective to align the target speech tokens.

\subsection{Boundary-aware Speech Prompt}
\label{cif-prompt}
\begin{figure}[t]
    \centering
    \includegraphics[width=63mm]{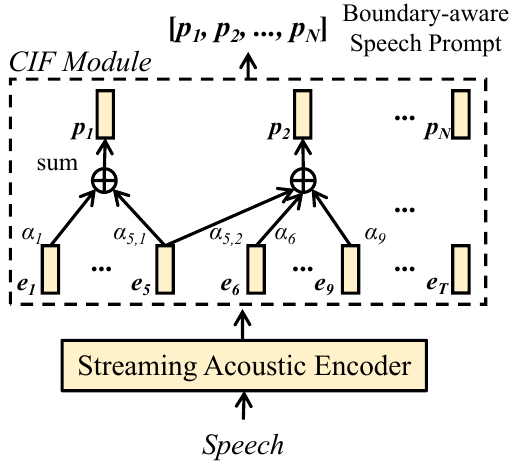}
    \vspace{-0.25cm}
    \caption{Illustration of the boundary-aware speech prompt extraction using the CIF. $\bm{\oplus}$ denotes addition. }
    \vspace{-0.2cm}
    \label{cif}
\end{figure}
The challenges faced by SimulS2S-LLM mainly stem from the mismatch between offline training and simultaneous inference. This paper does not focus on scenarios with extremely low latency (e.g., AL < 1~s in Simul-S2TT), as prior work \cite{deng2024lst} indicates these are unsuitable for offline-trained models and compromise translation quality due to the need for re-ordering.

Preliminary experiments showed that simply using down-sampling methods, such as stacking encoder outputs \cite{10447605, 10445874, ma2024embarrassingly} to obtain speech prompts, leads to poor performance during simultaneous inference after offline training.
However, inspired by the fact that the test-time wait-k strategy works well in text-based simultaneous machine translation \cite{gu-etal-2017-learning, Ma2018STACLST}, this paper proposes that the key to unlocking simultaneous inference for offline-trained speech LLMs is extracting boundary-aware speech prompts, which makes the system closer to the text-based scenario. Simple down-sampling during simultaneous decoding, where only partial speech prompts are available, ignores word boundary information and prevents the model from making correct predictions.

SimulS2S-LLM obtains the boundary-aware speech prompts using the CIF\footnote{The visualisation of CIF alignment refers to the supplementary materials of \citet{10572334}} mechanism \cite{9054250}, a non-autoregressive method that jointly learns alignments and high-level representations. To be more specific,
a scalar weight $\alpha_t$ is learned for each encoder output frame $\bm{e}_t$, and the boundary-aware speech prompts $\bm{p}_i$ are obtained via weighted addition. Following \cite{deng2024lst}, this paper simply uses the last dimension of $\bm{e}_t$ as the raw scalar attention value $\alpha_t$ to avoid additional parameters: $\alpha_t = {\rm sigmoid}({e_{t,d}})$, where $d$ is the dimension size of $\bm{e}_t$. The weights $\alpha_t$ are accumulated from left to right (i.e., to support streaming) until the sum exceeds a threshold of 1.0. Once the threshold is reached, the current weight $\alpha_t$ is split into two parts $\alpha_{t,1}$ and $\alpha_{t,2}$: $\alpha_{t,1}$ ensures the accumulation of exactly 1.0, while $\alpha_{t,2}$ is used for the next integration. For instance, as shown in Fig.~\ref{cif}, if the threshold 1.0 is reached at $t=5$, the boundary-aware speech prompts at the $1$-st step can be obtained via: $\bm{p}_1 = \sum_{j=1}^{4}\alpha_{j}\cdot\bm{e}_{j,1:d-1}+\alpha_{5,1}\cdot\bm{e}_{5,1:d-1}$. The $\bm{p}_i$ will be mapped to the same dimension size as the LLM embedding size before being fed in. The accumulation is then reset to zero and conducted incrementally.
To learn the CIF alignment, a quantity loss $\mathcal{L}_{\rm qua}=|\sum_{j=1}^T\alpha_j - N|$ is calculated during training, guiding accumulated weights to align with the source text length (N).

\subsection{Offline Training of SimulS2S-LLM}
SimulS2S-LLM uses a two-stage training strategy. 
The first stage of training corresponds to the speech-to-text translation task. An off-the-shelf text-based LLM is used and kept fixed. The encoder and CIF are optimised under the supervision of the cross-entropy loss function as shown in Fig.~\ref{SimulS2S-LLM}, where the target-language text tokens ($y_1, \cdots, [eos]$)
is used as the training target. 
In addition, the quantity loss is also considered:
 \begin{equation}
    \mathcal{L}_{\rm Train}^{\rm First}=\mathcal{L}_{\rm CE} + \gamma \mathcal{L}_{\rm qua} \label{obj-train-1}
\end{equation}

Note the entire speech prompt is pre-pended to the input text embedding sequence  ($\bm{z}_0 \cdots \bm{z}_N$), enabling offline training.
In addition, the template instructions that determine the translation task, e.g. ``Translate the French text into English", are used in both the first and second-stage training.

In the second stage of training, only the layer-wise weights $\beta_i$ and speech generator are updated under the supervision of the CTC loss, where the speech semantic tokens are used as the target.

\begin{figure*}[t]
    \centering
    \includegraphics[width=160mm]{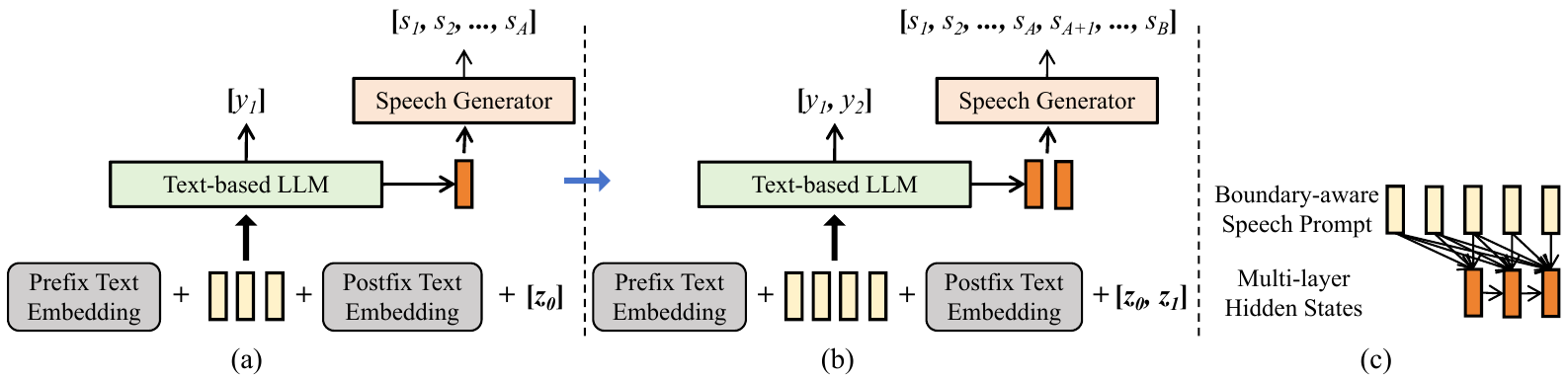}
    \caption{Illustration of the simultaneous inference of offline-trained SimulS2S-LLM (with wait-3 as an example) (a) the generation of the $1$-st hidden state and corresponding speech tokens $t_i$; (b) $2$-nd generation; (c) overall illustration of the hidden state generation order according to the speech prompt (always wait 3 more steps here).}
    \label{decode}
\end{figure*}

\subsection{Simultaneous Inference of SimulS2S-LLM}
\label{sec:simul-infer}
SimulS2S-LLM uses the wait-k strategy during testing to achieve simultaneous inference. The CIF module, along with the streaming encoder, extracts the speech prompts online, whose length is at the text token level. Therefore, the inference process is determined and driven by CIF. When $k=3$ in wait-k, as shown in Fig.~\ref{decode}, the hidden states corresponding to the translation generated by the LLM are always two steps behind the speech prompt.
For example, in Fig.~\ref{decode}, a speech prompt of length 3 corresponds to the generation of the first LLM token (Fig.~\ref{decode}(a)), while a speech prompt of length 4 corresponds to the generation of the second LLM token (Fig.~\ref{decode}(b)).
Note when a new speech prompt is obtained from newly received speech, past keys and values including positional information need to be updated accordingly before LLM generation.
Once the entire speech input is loaded, the LLM is no longer constrained by the speech prompt length and completes the prediction auto-regressively, making use of tail beam search \cite{Ma2018STACLST}.

Whenever a new hidden state is generated, it is fed into the speech generator, where it undergoes up-sampling (e.g., by a factor of sampling rate $U$) before being passed into a causal Transformer layer. The speech generator then outputs new CTC logits. To mitigate the independence assumption of CTC, a speech token-based n-gram LM is built to assist the CTC frame-synchronous decoding via shallow fusion. To expand the search space without introducing additional latency, an incremental beam search is designed. Specifically, within the range of new CTC logits (i.e., of length $U$, which is the up-sampling rate)), decoding is performed frame by frame using beam search. After decoding the final frame, only the highest-probability hypothesis is retained, while other hypotheses are pruned. For example, the predicted speech tokens ($s_{1} \cdots s_{A}$) in Fig.~\ref{decode}(a) are the prefixes of the predicted speech tokens in Fig.~\ref{decode}(b).
Note this pruning is no longer needed once the input speech has been fully loaded.

This paper uses a chunk-based mask operation to implement the streaming Transformer encoder. Therefore, at inference, the speech input is loaded chunk by chunk, and CIF continues to accumulate $\alpha_t$ based on the newly read speech chunk, dynamically generating new speech prompts. Before the entire speech input is read, the number of new LLM tokens generated for each new speech chunk is still determined by wait-k, i.e. the total length of LLM tokens remains $k$ shorter than the latest speech prompt length.
If multiple LLM tokens can be generated within a single speech chunk, beam search is used to expand the search space. After generating the last LLM token in each chunk, only the highest-probability hypothesis is retained to implement pruning, avoiding additional latency.

\begin{algorithm}[t]
\caption{SimulS2S-LLM Inference} 
\footnotesize 
\begin{algorithmic}[1]
\Require {$\textbf{E}_{:(n+1)*c}, \bm{y}, L_\text{max}, K,$ \textit{Final}}
\Ensure {$\bm{s}^\text{gen}$}
\State $L_\text{prev} \gets {\rm len}(\bm{y})$: Get the previous token $\bm{y}$ length $L_\text{prev}$
\State $L_{p}, \textbf{p} \gets {\rm CIF(\textbf{E}_{:(n+1)*c})}$: Get speech prompt $\textbf{p}$ and its length $L_{p}$ with the input chunks of speech $\textbf{E}_{0:(n+1)*c}$
\If {\textit{Final}}
    \State $L_\text{gen} \gets L_\text{max}$: Set the new token number $L_\text{gen}$ to the max length $L_\text{max}$ if the input is the final complete one
\Else
    \State $L_\text{gen} \gets (L_{p} - L_\text{prev} -K +1)$: Guided by wait-k
\EndIf 
\If {$L_\text{gen}<=0$}
    \State \Return
\EndIf
\State $\bm{y}^\text{gen}, \textbf{h}^\text{gen} \gets {\rm LLM}(\textbf{p}, L_\text{gen})$: New tokens $\bm{y}^\text{gen}$ and hidden states $\textbf{h}^\text{gen}$ based on \textbf{p} and length constraint $L_\text{gen}$
\State $\bm{s}^\text{gen} \gets {\rm Speech\text{-}Generator}(\textbf{h}^\text{gen})$: New speech tokens
\State \Return $\bm{s}^\text{gen}$
\end{algorithmic}
\end{algorithm}

The detailed procedure for this simultaneous inference is shown in Algorithm 1, 
where the inputs are the streaming speech encoder output $\textbf{E}_{:(n+1)*c}$ (with a chunk size of $c$), previously predicted tokens $\bm{y}$,
the maximum generation length $ L_\text{max}$, the wait-k $K$ steps, and whether the input speech is now a complete utterance, denoted \textit{Final}. Then the newly predicted speech semantic tokens $\bm{s}^\text{gen}$ ($s_{1} \cdots s_{A}\cdots$) will be returned. Note that the functions ${\rm LLM}$ and ${\rm Speech\text{-}Generator}$ in Algorithm 1 have recorded the past keys and values in a cache, so only the speech content of the current chunk is needed to complete the generation.

\section{Experimental Setup}
\label{setup}
\subsection{Dataset}
Experiments were conducted on CVSS-C data \cite{jia2022cvss}, which is a large-scale speech-to-speech translation data created from the CoVoST 2 \cite{wang2021covost} speech-to-text translation dataset with synthesised target speech. SimulS2S-LLM was evaluated on Spanish-English (Es-En), French-English (Fr-En), and German-English (De-En) pairs. Additional details about the data are provided in Appendix~\ref{sec:appendix:data}.

\subsection{Model Descriptions}

Speech semantic tokens were extracted from the target speech using mHuBERT \cite{DBLP:conf/interspeech/PopuriCWPAGHL22}. Based on the training set, target speech token-based 4-gram LMs were obtained using KenLM toolkit, which was incorporated into CTC decoding with 0.5 weight. The raw speech waveform was used as input.  For Es-En and Fr-En, BLOOMZ-7B1 \citep{le2023bloom} was used as the text-based LLM and kept fixed all the time. 
For De-En, Llama3-8B \cite{dubey2024llama} was used as BLOOMZ underperforms in German.
A pre-trained unit-based HiFi-GAN vocoder
\cite{kong2020hifi} was used to synthesise speech. To achieve streaming speech generation, partially predicted speech tokens are directly sent to the vocoder. The resulting audio signal is used as the prefix for the next prediction and will no longer be modified.

All models built in this paper used the same streaming Transformer encoder, fine-tuned from the "xlsr\_53\_56k" model provided by Fairseq \cite{ott2019fairseq}, with a chunk-based masking operation. 
The chunk size was set to 32, corresponding to a theoretical average latency of 320 ms. More details can be found in Appendix~\ref{hyper}.


\paragraph{SimulS2S-LLM}
In addition to the encoder, as mentioned in Sec.~\ref{cif-prompt}, the CIF module only involves a fully-connected (FC) layer to map the speech prompt dimension to the LLM embedding dimension, i.e. 4096. 
The speech generator consists of 8 causal Transformer layers (1024 attention dimension, 2048 feed-forward dimension, and 8 heads), which use subsequent masks to avoid seeing future information.
The up-sampling rate $U$ was set to 25. The beam size of the incremental beam search was set to 10.

\paragraph{Boundary-unaware SimulS2S-LLM}
A fixed down-sampling method \citep{10447605} was implemented to extract a boundary-unaware speech prompt for comparison with SimulS2S-LLM. This model, referred to as boundary-unaware SimulS2S-LLM, serves as the baseline model. Following \cite{10447605}, considering the frame stride of the encoder was 20~ms, every 16 consecutive acoustic encoder output frames were stacked to achieve down-sampling. Then, an additional FC layer was applied to map the stacked encoder outputs to the LLM embedding dimension (i.e., 4096) before feeding them into the LLM. This model also used the wait-k policy for inference, with every fixed 16 encoder outputs used as one step.

\paragraph{StreamSpeech}
StreamSpeech \cite{zhang-etal-2024-streamspeech} is a recent Simul-S2ST model that has achieved state-of-the-art (SOTA) results, with the same speech tokens and vocoder used as in SimulS2S-LLM. Note that it is not an LLM-based approach and is used to provide a benchmark result.

\subsection{Metrics}
Experiments were implemented based on the ESPnet-ST \cite{inaguma-etal-2020-espnet}.
SimulEval \cite{ma-etal-2020-simuleval} was used to evaluate the models.
\vspace{-2mm}
\paragraph{ASR-BLEU}
For Simul-S2ST, the ASR-BLEU 
toolkit\footnote{\url{https://github.com/facebookresearch/fairseq/tree/ust/examples/speech_to_speech/asr_bleu}} was used to evaluate the translation quality, which transcribes the synthesised speech into text before calculating SacreBLEU \cite{DBLP:conf/wmt/Post18} with the reference text. 
\vspace{-2mm}
\paragraph{ATD}
Following the Simuleval example\footnote{\url{https://github.com/facebookresearch/SimulEval/tree/main/examples/speech_to_speech}},
the average token delay (ATD) 
\cite{kano2022average} was used to measure the speech generation latency.
ATD refers to the average delay between output sub-segments and corresponding input sub-segments. 

Text output Simul-S2TT was also evaluated, and the translation quality was measured using SacreBLEU.
The speech version of the word-level Average Lagging (AL) \cite{Ma2018STACLST, Ma2020SimulMTTS} was used to measure latency.

\begin{figure}[t!]
	\centering
	\begin{minipage}{\linewidth}
		\centering
		\includegraphics[width=1.0\linewidth]{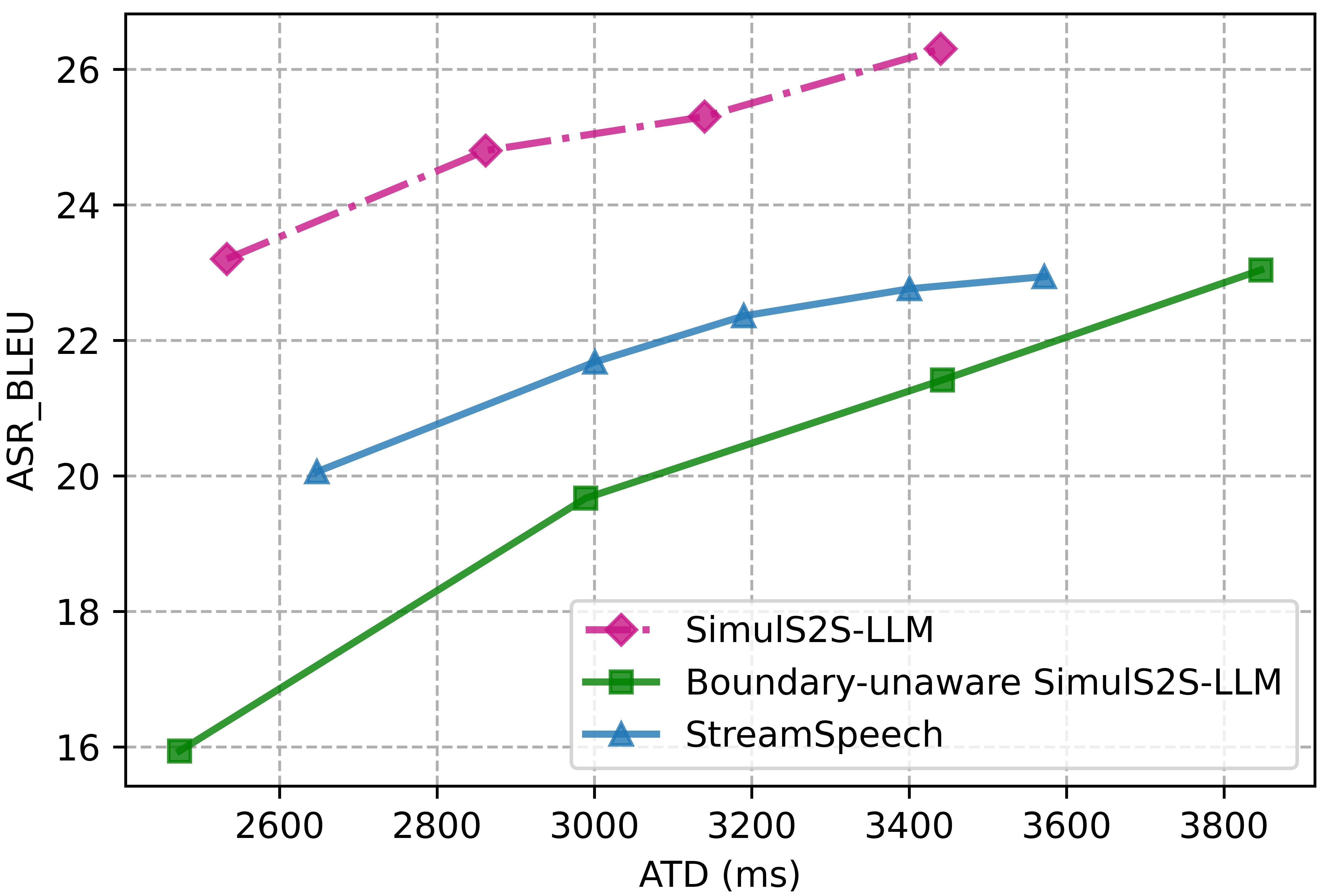}
        \vspace{-5mm}
		\caption*{Simul-S2ST Es-En}
		\label{chutian1}
	\end{minipage}
	
	\vspace{2mm} 

	\begin{minipage}{\linewidth}
		\centering
		\includegraphics[width=1.0\linewidth]{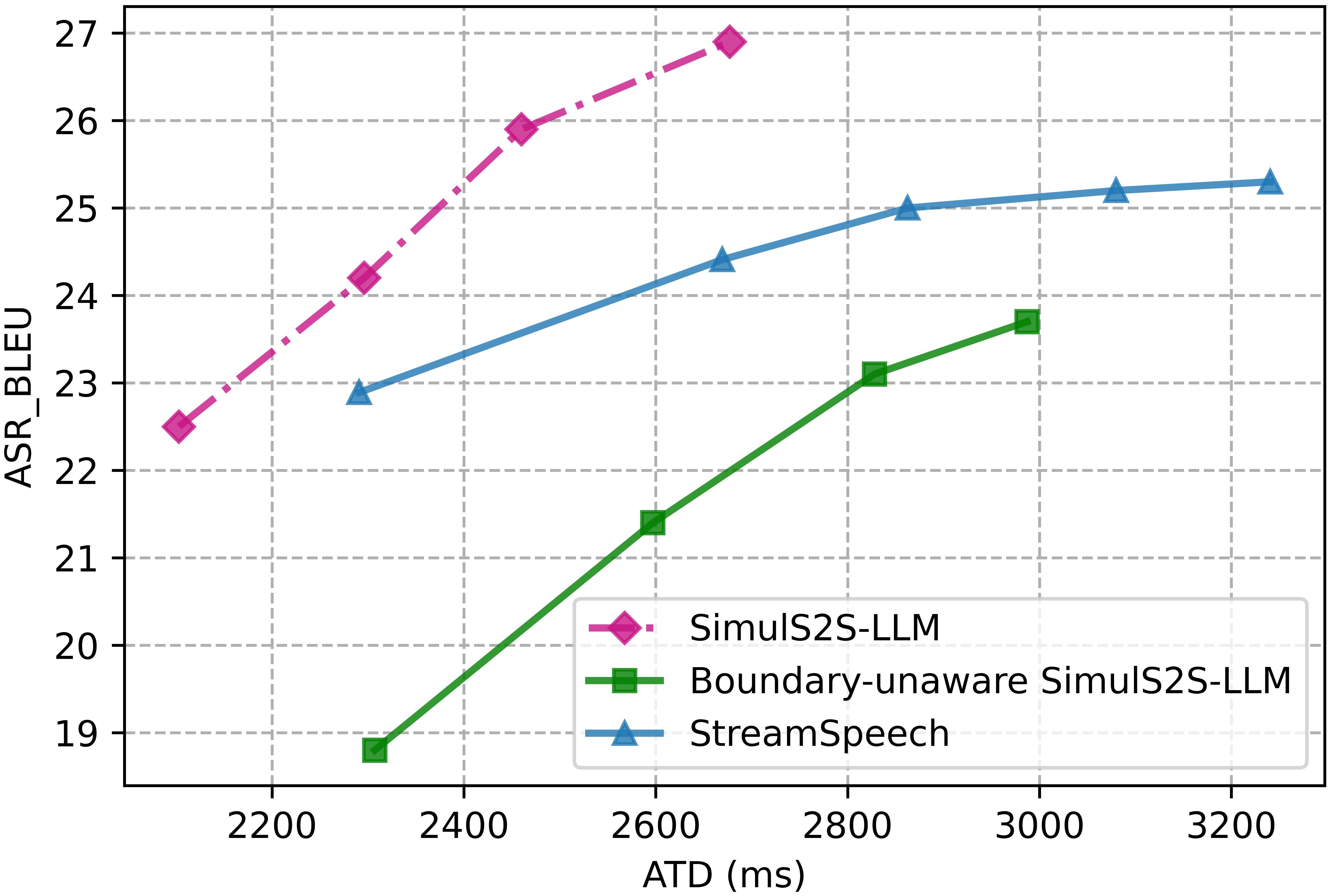}
        \vspace{-5mm}
		\caption*{Simul-S2ST Fr-En}
		\label{chutian2}
	\end{minipage}
	
	\vspace{2mm}

	\begin{minipage}{\linewidth}
		\centering
		\includegraphics[width=1.0\linewidth]{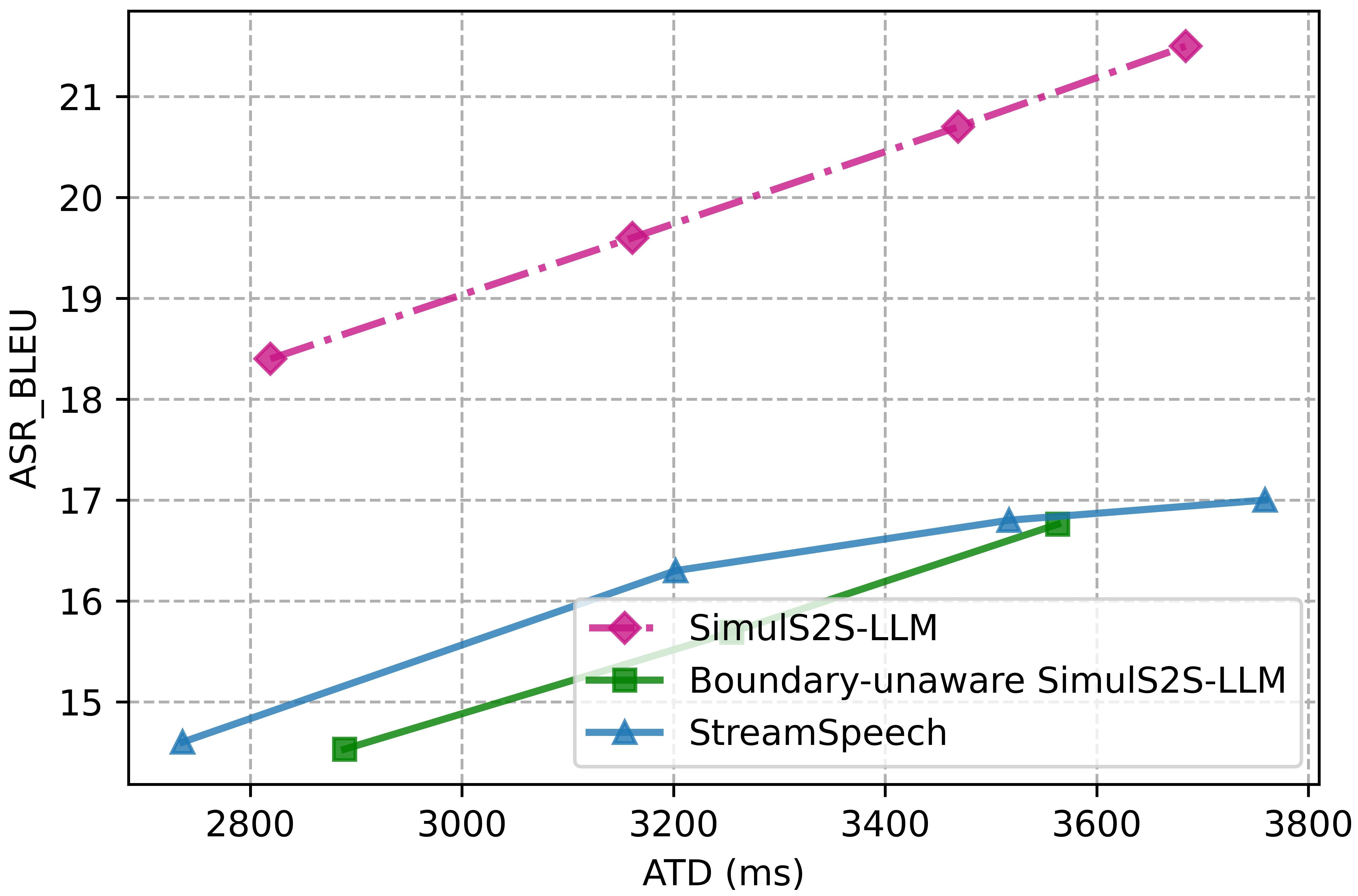}
        \vspace{-5mm}
		\caption*{Simul-S2ST De-En}
		\label{chutian3}
	\end{minipage}
    \vspace{-0.2cm}
	\caption{Simul-S2ST quality-latency trade-off curves on CVSS-C Es-En, Fr-En, and De-En test sets. The x-axis represents the latency, measured by ATD, and the y-axis represents the speech translation quality, measured by ASR-BLEU. Note that the y-axis scales can be different on different sub-figures.}
	\label{main-s2s}
    \vspace{-0.2cm}
\end{figure}

\section{Experimental Results}
This section compares the proposed SimulS2S-LLM with existing methods, such as StreamSpeech, as well as boundary-unaware SimulS2S-LLM. Ablation studies were conducted to evaluate the effectiveness of boundary-aware speech prompts,  utilising multi-layer hidden states and speech token generation.
Note, in order to ensure high translation quality, SimulS2S-LLM does not target scenarios requiring extremely low latency.

\subsection{Simul-S2ST Results}

\begin{table}[t] 
	\centering
	\setlength{\tabcolsep}{1.1mm}
	\renewcommand\arraystretch{1.0}
	\begin{tabular}{ c l  c c c}
		\Xhline{3\arrayrulewidth}
		\multicolumn{2}{c}{S2ST Models}&Es-En&Fr-En&De-En\\
		\hline
            \multicolumn{2}{l}{\emph{Offline}} \\
		&S2UT &18.53&22.23&--\\
		&Translatotron &8.72&16.96&--\\
		&Translatotron 2&22.93&26.07&16.91\\
		&DASpeech &21.37&25.03&16.14\\
		&UnitY &24.95&27.77&18.74\\
		&Offline StreamSpeech&27.25&28.45&20.93\\
		\hline
        \multicolumn{2}{l}{\emph{Streaming}} \\
            &StreamSpeech&22.94&25.30&17.0\\
		&SimulS2S-LLM &26.33&26.93&21.5\\
		\Xhline{3\arrayrulewidth}
	\end{tabular}
	\vspace{-0.15cm}
	\caption{ASR-BLEU ($\uparrow$) results on the CVSS-C data for different models, including S2UT \cite{lee-etal-2022-direct}, Translatotron \cite{DBLP:conf/interspeech/JiaWBMJCW19}, Translatotron 2\cite{DBLP:conf/icml/JiaRRP22}, DASpeech \cite{DBLP:conf/nips/FangZ023a}, UnitY \cite{DBLP:conf/acl/InagumaPKCWC00023}. The SimulS2S-LLM results correspond to the last points in Fig.~\ref{main-s2s}. Note the comparisons are not well-controlled. The published benchmark results are reproduced from \citet{zhang-etal-2024-streamspeech} on CVSS-C.}
	\label{benchmark}
	\vspace{-0.1cm}
\end{table}

Figure~\ref{main-s2s} shows the Simul-S2ST results on CVSS-C Es-En, Fr-En, and De-En data, with the ASR-BLEU scores plotted against ATD.
Although SimulS2S-LLM was trained offline, it still clearly outperforms the strong StreamSpeech models with simultaneous inference.
For example, on the Es-En test set, SimulS2S-LLM outperformed StreamSpeech by approximately 4 ASR-BLEU points while maintaining the same latency. This demonstrates that SimulS2S-LLM can effectively leverage the strong LLM generation capabilities in a streaming manner. Previous work has shown that text-based LLMs can be extended to speech with strong performance across a wide range of tasks, such as translation and question answering \cite{tang2024salmonn, chu2023qwen}. SimulS2S-LLM further unlocks simultaneous inference while potentially retaining these emergent abilities by following the same training paradigm. Additionally, the boundary-unaware version of SimulS2S-LLM failed to achieve such strong performance, in line with our expectation that learning boundary-aware speech prompts can unlock the simultaneous inference abilities of offline-trained speech LLMs. With a boundary-aware speech prompt, SimulS2S-LLM shares more similarities with
text-based simultaneous translation, where test-time wait-k is commonly used. 
Henceence, the extensive comparisons with both StreamSpeech and the boundary-unaware SimulS2S-LLM provide strong evidence for the superiority of our method.
The numerical results in Fig.~\ref{main-s2s} are given in Appendix~\ref{numeric} and extended computation-aware results are shown in Appendix~\ref{numeric_ca}.

Table~\ref{benchmark} compares streaming SimulS2S-LLM with published speech-to-speech translation (S2ST) results from the literature on CVSS-C, showing that SimulS2S-LLM achieves competitive performance as a streaming model.
In the streaming scenario, the results for SimulS2S-LLM and StreamSpeech are represented by the last points in Fig.~\ref{main-s2s}.

Appendix~\ref{compare_sota} compares with SOTA models like SeamlessStreaming \cite{barrault2023seamless} which uses 9,300 hours of speech-to-speech data in contrast to between 
69.5 and 174 hours of speech-to-speech data for individual language pairs used here. 

\begin{figure}[t]
    \centering
    \includegraphics[width=68mm]{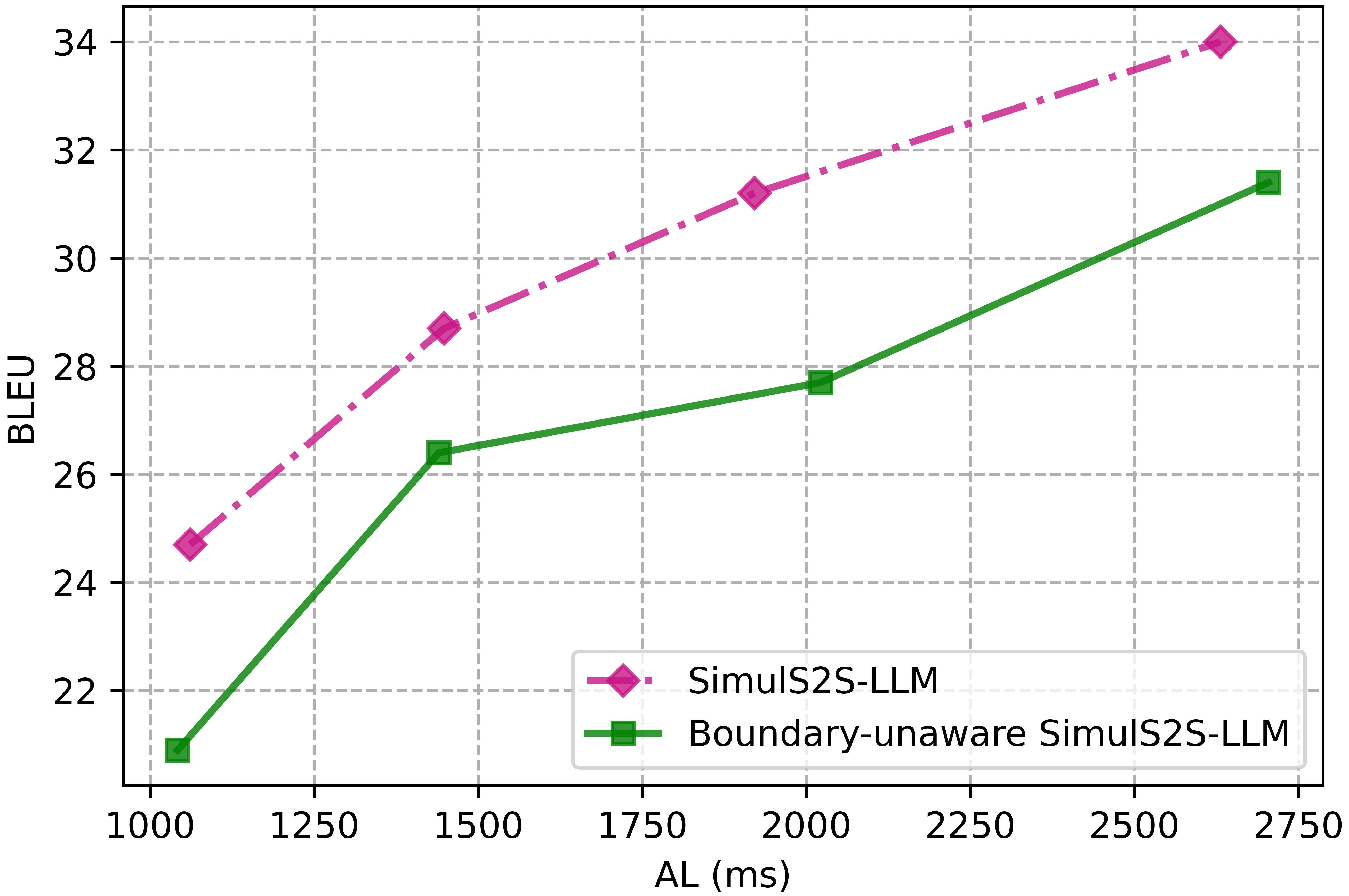}
    \vspace{-0.1cm}
    \caption{Simul-S2TT quality-latency trade-off curves on CVSS-C Es-En test set. The x-axis represents the latency, measured by AL, and the y-axis represents the speech translation quality, measured by BLEU.}
    \vspace{-0.2cm}
    \label{main-s2t}
\end{figure}

\subsection{Ablation on Speech Prompt Type for Simul-S2TT}
This sub-section compares SimulS2S-LLM and boundary-unaware SimulS2S-LLM on the Simul-S2TT task. Since the only difference between them is the speech prompt used, this comparison can effectively evaluate the importance of boundary-aware speech prompts in unlocking simultaneous inference. As shown in Fig.~\ref{main-s2t}, SimulS2S-LLM consistently outperformed the boundary-unaware one. For example, with similar latency, the 
boundary-aware SimulS2S-LLM was about 4 BLEU points higher. Hence, the experimental results on both the Simul-S2ST and Simul-S2TT tasks demonstrate the importance of using boundary-aware speech prompts for offline-trained speech LLMs. The numerical results in Fig.~\ref{main-s2t} are given in Appendix~\ref{numeric-s2t}


Moreover, by comparing the differences in translation quality between SimulS2S-LLM and boundary-unaware SimulS2S-LLM in Fig.~\ref{main-s2s} and Fig.~\ref{main-s2t}, it can be observed that the gap is similar, with the ASR-BLEU and BLEU values both differing by around 4 points at similar latencies. Therefore, the main reason for the poorer performance of boundary-unaware SimulS2S-LLM is the prediction error of the LLM, which is in line with expectations as they use the same speech generator.

\subsection{Ablation on Multi-layer Hidden States}
An ablation study was conducted to evaluate the effectiveness of using multiple layers of LLM hidden states. Fig.~\ref{multi-layer} shows that leveraging LLM multi-layer hidden states is more beneficial for predicting speech tokens, causing about one ASR-BLEU point improvement. The final hidden layer focuses on semantic information \cite{DBLP:conf/interspeech/ChangYFM023}, which is favourable for text token prediction, whereas multiple hidden states capture richer information. Moreover, due to the mismatch caused by teacher forcing in training, using multiple hidden states seems to be more robust.

\subsection{Ablation on Speech Token Generation}
This sub-section compares the use of n-gram and greedy search for predicting discrete speech tokens. As shown in Table~\ref{n-gram-study}, although discrete speech tokens are more challenging to predict than text units, n-gram LMs based on discrete speech tokens can still assist CTC in making more accurate predictions, thus improving the translated speech quality.

\begin{figure}[t]
    \centering
    \includegraphics[width=70mm]{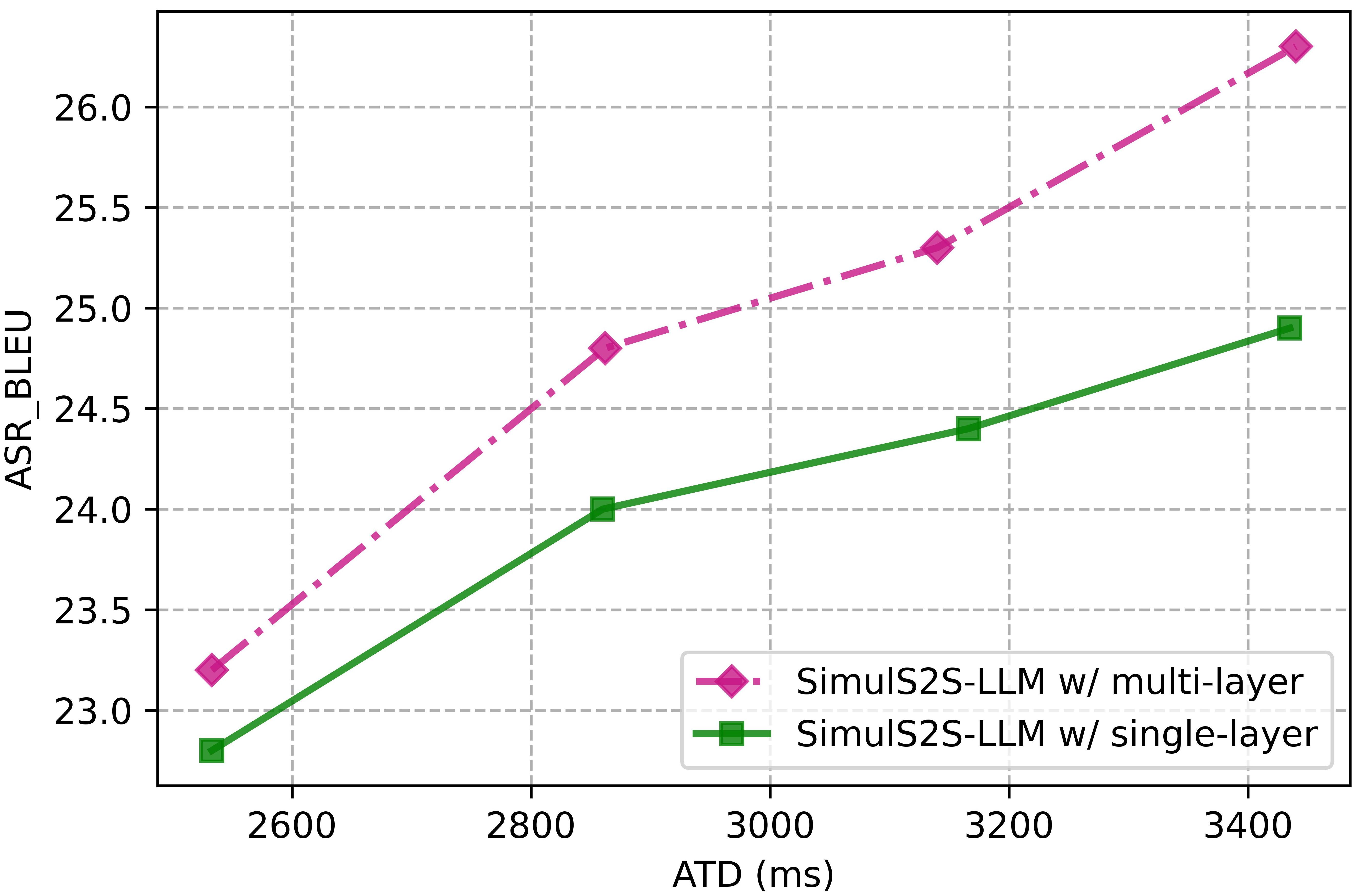}
    \vspace{-0.1cm}
    \caption{Simul-S2ST Es-En results for SimulS2S-LLM with single or multi-layer LLM hidden states.}
    \label{multi-layer}
\end{figure}

\begin{table}[t] 
	\centering
	\setlength{\tabcolsep}{0.1mm}
	\begin{tabular}{ l| c| c }
		\Xhline{3\arrayrulewidth}
        {Models}&{ASR-BLEU}&{ATD (ms)}\\
		\hline
		SimulS2S-LLM w/ n-gram &26.3&3440\\
        SimulS2S-LLM w/ greedy &24.7&3439\\
		\Xhline{3\arrayrulewidth}
	\end{tabular}
	\vspace{-0.1cm}
	\caption{ASR-BLEU ($\uparrow$) results on the CVSS-C Es-En data for different speech token generation methods.}
	\label{n-gram-study}
	\vspace{-0.1cm}
\end{table}

\subsection{Speech Evaluation with BLASER 2.0}
This section further uses BLASER 2.0 \cite{dale-costa-jussa-2024-blaser} to evaluate the generated speech quality of SimulS2S-LLM. BLASER 2.0 includes three scores: Unsupervised (0–1), QE (1–5), and Ref (1–5). The Unsupervised version computes cosine similarity between sentence-level embeddings without supervision, while QE and Ref are supervised models trained to predict human ratings, with Ref additionally requiring reference target speech.

Table~\ref{BLASER-2.0-analysis} results show that as a streaming model, SimulS2S-LLM gave higher BLASER 2.0 scores compared to the offline StreamSpeech and UnitY models on the CVSS-C benchmark data, demonstrating superior translation and speech quality.

\begin{table}[t] 
	\centering
	\setlength{\tabcolsep}{1.0mm}
	\renewcommand\arraystretch{1.0}
	\begin{tabular}{ l | c| c| c}
		\Xhline{3\arrayrulewidth}
		{S2ST Models}&Unsupervised&QE&Ref\\
		\hline
        Offline UnitY&0.51&3.33&3.26\\
        Offline StreamSpeech&0.52&3.37&3.35\\
		SimulS2S-LLM &0.72&3.72&3.59\\
		\Xhline{3\arrayrulewidth}
	\end{tabular}
	\caption{BLASER 2.0 ($\uparrow$) results on the CVSS-C Es-En data. The published results of offline UnitY and StreamSpeech are from \citet{zhang-etal-2024-streamspeech}.}
	\label{BLASER-2.0-analysis}
\end{table}

\section{Conclusions}

This paper proposes SimulS2S-LLM, the first work to extend LLMs to Simul-S2ST while
avoiding being constrained to specific streaming tasks via offline training.
SimulS2S-LLM uses a test-time wait-k policy to guide the simultaneous inference.
To alleviate offline training and simultaneous inference mismatch, SimulS2S-LLM extracts boundary-aware speech prompts based on CIF. 
To generate high-quality speech in streaming, multi-layer LLM hidden states are used by a causal Transformer-based speech generator to predict discrete speech tokens. To enhance this prediction process, an incremental beam search is designed to expand the search space of speech tokens without introducing additional latency, while a speech token-based n-gram LM is also incorporated. Experiments show that SimulS2S-LLM gives a better quality-latency trade-off than existing Simul-S2ST methods.

\section*{Limitations}

This paper has the following limitations.
\begin{enumerate}
    \item SimulS2S-LLM relies on off-the-shelf text-based LLMs, meaning that the performance is inherently constrained by the capabilities of the available text-based models. Due to limited computing resources, this paper focuses on using 7B/8B LLMs, as larger models are beyond our computational capacity. Additionally, SimulS2S-LLM is restricted to open-source LLMs and cannot use closed-source models like GPT-4.

\item This paper does not focus on scenarios with extremely low latency (e.g., AL < 1~s in Simul-S2TT), as prior work \cite{deng2024lst} indicates these are unsuitable for offline-trained models and compromise translation quality due to the need for re-ordering. Moreover, using LLMs increases the computational load, which leads to higher latency when considering computation time. This is a common challenge faced by the community, and significant research development is needed to accelerate LLM inference speed. As such, this aspect is beyond the scope of this paper and is left as future work.
In addition, since this is the first work to apply LLM to the Simul-S2ST task, we couldn't find an LLM-based method to compare with SimulS2S-LLM on the Simul-S2ST task.

\item As mentioned in Section~\ref{sec:simul-infer}, during the simultaneous inference of the proposed SimulS2S-LLM, when a new speech chunk is read in, the past keys and values need to be updated before LLM generation. However, according to the analysis in Appendix~\ref{numeric_ca}, the time consumed by this process should not be significant as it can be performed in parallel.
This paper has also not evaluated SimulS2S-LLM on long-form Simul-S2ST due to the lack of data.

\item Due to limitations in training data and computing resources, we were unable to train our SimulS2S-LLM as extensively as some foundation models like SeamlessStreaming \cite{barrault2023seamless} which uses 9,300 hours of speech-speech data. However, we conducted comprehensive experiments across three language pairs and CVSS-C is the most widely used speech-to-speech translation data set, even though the individual language pairs have only between 69.5 and 174 hours of speech-to-speech data. In addition, if more speech-to-speech translation data is used, the speech generation performance of SimulS2S-LLM can be expected to greatly improve.

\item This paper evaluates SimulS2S-LLM on three European language pairs, each in a single translation direction (i.e., Es-En, Fr-En, and De-En). While we believe the technique can be extended to other languages, including non-European ones, and additional translation directions, its performance in these cases remains unverified and is left for future work. In addition, simultaneous translation varies in difficulty for different language pairs due to the extent of re-ordering, so achieving Simul-S2ST for certain language pairs can be challenging.

\item This paper evaluates SimulS2S-LLM only on simultaneous speech translation tasks, including Simul-S2TT and Simul-S2ST. Although it claims that SimulS2S-LLM avoids constraining speech LLMs to specific streaming tasks through offline training, it does not directly evaluate its performance on other simultaneous inference tasks with speech as the input or on offline inference tasks. This is because preserving the zero-shot task capabilities of speech LLMs is not the main focus of this paper and has already been extensively studied in prior work.
\end{enumerate}
\section*{Ethics Statement}
Deep learning systems are data-hungry, and without sufficient data, it is difficult to achieve promising model performance.  For under-resourced languages or domains, this issue will be even more severe.
This can lead to a poor user experience for minority groups, resulting in their views being underrepresented or misunderstood.
The SimulS2S-LLM technique proposed in this paper can alleviate this issue by translating low-resource data into high-resource data in a low-latency manner, enabling the model to better handle the task.

\bibliography{ref, custom}

\clearpage
\appendix


\begin{table*}[t!] 
  \centering
  \setlength{\tabcolsep}{1.5mm}
  \renewcommand\arraystretch{0.97}
  \begin{tabular}{l | c |c |c| c  }
    \Xhline{3\arrayrulewidth}
     Simul-S2ST Models&ASR-BLEU&ATD&StartOffset&EndOffset\\
    \hline
    SimulS2S-LLM (k=5)&23.2&2533&3109&1722\\
    SimulS2S-LLM (k=6)&24.8&2862&3484&1788\\
    SimulS2S-LLM (k=7)&25.3&3140&3828&2055\\
    SimulS2S-LLM (k=8)&26.3&3440&4191&2209\\
    \hline
    Boundary-unaware SimulS2S-LLM (k=10)&17.7&2649&3810&2885\\
    Boundary-unaware SimulS2S-LLM (k=11)&19.7&2989&3811&2849\\
    Boundary-unaware SimulS2S-LLM (k=13)&21.4&3442&4374&2904\\
    Boundary-unaware SimulS2S-LLM (k=15)&23.0&3847&4879&3014\\
     \Xhline{3\arrayrulewidth}
  \end{tabular}
  \caption{Numerical results of SimulS2S-LLM on CVSS-C Es-En corresponding to Fig.~\ref{main-s2s}.}
  \label{tab:numeric-es}
\end{table*}

\begin{table*}[t!] 
  \centering
  \setlength{\tabcolsep}{1.5mm}
  \renewcommand\arraystretch{0.97}
  \begin{tabular}{l | c |c |c| c  }
    \Xhline{3\arrayrulewidth}
     Simul-S2ST Models&ASR-BLEU&ATD&StartOffset&EndOffset\\
    \hline
    SimulS2S-LLM (k=5)&22.5&2103&2659&1659\\
    SimulS2S-LLM (k=6)&24.2&2296&2914&1763\\
    SimulS2S-LLM (k=7)&25.9&2460&3136&1857\\
    SimulS2S-LLM (k=8)&26.9&2677&3378&1914\\
    \hline
    Boundary-unaware SimulS2S-LLM (k=9)&18.8&2307&3025&2309\\
    Boundary-unaware SimulS2S-LLM (k=11)&21.4
    &2597&3447&2363\\
    Boundary-unaware SimulS2S-LLM (k=13)&23.1&2829&3770&2421\\
    Boundary-unaware SimulS2S-LLM (k=15)&23.7&2987&4014&2518\\
     \Xhline{3\arrayrulewidth}
  \end{tabular}
  \caption{Numerical results of SimulS2S-LLM on CVSS-C Fr-En corresponding to Fig.~\ref{main-s2s}.}
  \label{tab:numeric-fr}
\end{table*}

\begin{table}[t!] 
\centering
\setlength{\tabcolsep}{1.0mm}
\renewcommand\arraystretch{1.0}
\begin{tabular}{ l|c|c }
\Xhline{2\arrayrulewidth}
\hline
 &\multicolumn{2}{c}{CVSS-C Es-En} \\
\hline
Train set &\multicolumn{2}{c}{train}\\
\ \ -Duration&\multicolumn{2}{c}{69.5 hours}\\
\ \ -Sentences&\multicolumn{2}{c}{79K}\\
\cline{2-3}
Test sets&\multicolumn{1}{c|}{test} &dev\\
\ \ -Duration &\multicolumn{1}{c|}{12.4 hours}&{12.4 hours}\\
\ \ -Sentences&13K &13K\\
\hline
\hline
&\multicolumn{2}{c}{CVSS-C Fr-En}\\
\hline
Train set &\multicolumn{2}{c}{train}\\
\ \ -Duration&\multicolumn{2}{c}{174.0 hours}\\
\ \ -Sentences&\multicolumn{2}{c}{207K}\\
\cline{2-3}
    Test sets &test&dev\\
\ \ -Duration &13.3 hours& 13.0 hours\\
\ \ -Sentences&15K &15K\\
\hline
\hline
&\multicolumn{2}{c}{CVSS-C De-En}\\
\hline
Train set &\multicolumn{2}{c}{train}\\
\ \ -Duration&\multicolumn{2}{c}{112.4 hours}\\
\ \ -Sentences&\multicolumn{2}{c}{128K}\\
\cline{2-3}
    Test sets &test&dev\\
\ \ -Duration &12.1 hours& 12.5 hours\\
\ \ -Sentences&14K &14K\\
\Xhline{2\arrayrulewidth}
\end{tabular}
\caption{Statistics of datasets used in this paper}
\label{corpus}
\end{table}

\section{Data Statistics}
\label{sec:appendix:data}

The training and test data statistics are summarised in Table~\ref{corpus}. Data pre-processing followed ESPnet-ST recipes, including speed perturbation with factors of 0.9 and 1.1 during the first-stage training. Model training was conducted on two NVIDIA A100 GPUs, each with 80GB of memory. For CVSS-C Es-En, each epoch of first-stage training required approximately 3 hours, while second-stage training took about 20 minutes per epoch. For CVSS-C Fr-En, the first-stage training required around 9 hours per epoch, with the second stage taking approximately 1 hour per epoch. For CVSS-C De-En, the first-stage training required around 4 hours per epoch, with the second stage taking approximately 30 minutes per epoch.

\section{Hyper-parameters}
\label{hyper}
The hyper-parameters of the models we built are as follows, with other hyper-parameters following standard ESPnet-ST recipes.
 $\gamma$ in Eq.~\ref{obj-train-1} was set to 0.05. The beam size for LLM-based inference was set to 5, while the speech token-based incremental beam search used a beam size of 10. For SimulS2S-LLM on the Simul-S2ST task, the wait-$k$ policy was configured with $k \in \{5, 6, 7, 8\}$ on Es-En and Fr-En. For Simul-S2ST De-En, the wait-$k$ policy was configured with $k \in \{11, 13, 15, 17\}$, because Llama3-8B is English-centric, making German text token sequences longer than English. Hence, the learned CIF-based speech prompts become relatively longer and require larger $k$ values.
 For Simul-S2TT, $k$ was set to $k \in \{3, 4, 5, 7\}$ on Es-En. 
 $L_{max}$ for SimulS2S-LLM simultaneous inference is set to 0.15 times the length of the encoder output

\begin{table*}[t!] 
  \centering
  \setlength{\tabcolsep}{1.5mm}
  \renewcommand\arraystretch{0.97}
  \begin{tabular}{l | c |c |c| c  }
    \Xhline{3\arrayrulewidth}
     Simul-S2ST Models&ASR-BLEU&ATD&StartOffset&EndOffset\\
    \hline
    SimulS2S-LLM (k=11)&18.4&2819&3855&2182\\
    SimulS2S-LLM (k=13)&19.6&3161&4258&2435\\
    SimulS2S-LLM (k=15)&20.7&3469&4621&2636\\
    SimulS2S-LLM (k=17)&21.6&3684&4931&2838\\
    \hline
    Boundary-unaware SimulS2S-LLM (k=9)&12.4&2467&3176&2563\\
    Boundary-unaware SimulS2S-LLM (k=11)&14.5
    &2889&3757&2660\\
    Boundary-unaware SimulS2S-LLM (k=13)&15.7&3255&4270&2896\\
    Boundary-unaware SimulS2S-LLM (k=15)&16.8&3563&4697&3043\\
     \Xhline{3\arrayrulewidth}
  \end{tabular}
  \caption{Numerical results of SimulS2S-LLM on CVSS-C De-En corresponding to Fig.~\ref{main-s2s}.}
  \label{tab:numeric-de}
\end{table*}

\begin{table*}[t!] 
  \centering
  \setlength{\tabcolsep}{1.5mm}
  \renewcommand\arraystretch{0.97}
  \begin{tabular}{l | c |c |c| c  }
    \Xhline{3\arrayrulewidth}
     Simul-S2ST Models&ASR-BLEU&ATD\_CA&StartOffset\_CA&EndOffset\_CA\\
    \hline
    SimulS2S-LLM (k=5)&23.2&3114&3627&1722\\
    SimulS2S-LLM (k=6)&24.8&3462&4055&1788\\
    SimulS2S-LLM (k=7)&25.3&3816&4467&2055\\
    SimulS2S-LLM (k=8)&26.3&4239&4945&2209\\
    \hline
    Boundary-unaware SimulS2S-LLM (k=10)&17.7&3416&4486&2885\\
    Boundary-unaware SimulS2S-LLM (k=11)&19.7&3694&4292&2849\\
    Boundary-unaware SimulS2S-LLM (k=13)&21.4&4195&4939&2904\\
    Boundary-unaware SimulS2S-LLM (k=15)&23.0&4709&5573&3014\\
     \Xhline{3\arrayrulewidth}
  \end{tabular}
  \caption{Compution-aware results of SimulS2S-LLM on CVSS-C Es-En corresponding to Table~\ref{tab:numeric-es}.}
  \label{tab:numeric-es-ca}
\end{table*}

\begin{table*}[t!] 
  \centering
  \setlength{\tabcolsep}{1.5mm}
  \renewcommand\arraystretch{0.97}
  \begin{tabular}{l | c |c |c| c  }
    \Xhline{3\arrayrulewidth}
     Simul-S2ST Models&ASR-BLEU&ATD\_CA&StartOffset\_CA&EndOffset\_CA\\
    \hline
    SimulS2S-LLM (k=5)&22.5&2751&3201&1659\\
    SimulS2S-LLM (k=6)&24.2&2965&3502&1763\\
    SimulS2S-LLM (k=7)&25.9&3178&3784&1857\\
    SimulS2S-LLM (k=8)&26.9&3438&4077&1914\\
    \hline
    Boundary-unaware SimulS2S-LLM (k=9)&18.8&2993&3525&2309\\
    Boundary-unaware SimulS2S-LLM (k=11)&21.4
    &3347&4031&2363\\
    Boundary-unaware SimulS2S-LLM (k=13)&23.1&3652&4465&2421\\
    Boundary-unaware SimulS2S-LLM (k=15)&23.7&3893&4804&2518\\
     \Xhline{3\arrayrulewidth}
  \end{tabular}
  \caption{Compution-aware results of SimulS2S-LLM on CVSS-C Fr-En corresponding to Table~\ref{tab:numeric-fr}.}
  \label{tab:numeric-fr-ca}
\end{table*}

\begin{table*}[t!] 
  \centering
  \setlength{\tabcolsep}{1.5mm}
  \renewcommand\arraystretch{0.97}
  \begin{tabular}{l | c |c |c| c  }
    \Xhline{3\arrayrulewidth}
     Simul-S2ST Models&ASR-BLEU&ATD\_CA&StartOffset\_CA&EndOffset\_CA\\
    \hline
    SimulS2S-LLM (k=11)&18.4&3637&4589&2182\\
    SimulS2S-LLM (k=13)&19.6&4073&5095&2435\\
    SimulS2S-LLM (k=15)&20.7&4500&5575&2636\\
    SimulS2S-LLM (k=17)&21.6&4832&5987&2838\\
    \hline
    Boundary-unaware SimulS2S-LLM (k=9)&12.4&3127&3635&2563\\
    Boundary-unaware SimulS2S-LLM (k=11)&14.5
    &3528&4249&2660\\
    Boundary-unaware SimulS2S-LLM (k=13)&15.7&4007&4879&2896\\
    Boundary-unaware SimulS2S-LLM (k=15)&16.8&4432&5424&3043\\
     \Xhline{3\arrayrulewidth}
  \end{tabular}
  \caption{Compution-aware results of SimulS2S-LLM on CVSS-C De-En corresponding to Table~\ref{tab:numeric-fr}.}
  \label{tab:numeric-de-ca}
\end{table*}


\section{Numerical Values for Figure~\ref{main-s2s}}
\label{numeric}
The numerical values for Fig.~\ref{main-s2s} are provided in Tables~\ref{tab:numeric-es}, \ref{tab:numeric-fr}, and \ref{tab:numeric-de}. In addition to the ATD values displayed in Fig.~\ref{main-s2s}, the table includes the StartOffset and EndOffset metrics. StartOffset represents the delay before generating the first frame of the target speech, while EndOffset indicates the offset of the final frame of the target speech relative to the completion of the source speech. No matter which latency metric is used, the conclusion remains consistent.

\section{Compution-aware Latency Results}
\label{numeric_ca}
This section gives the latency results after considering the computation time, as shown in Table~\ref{tab:numeric-es-ca}, Table~\ref{tab:numeric-fr-ca}, and Table~\ref{tab:numeric-de-ca}. Note that, as mentioned in the Limitations section, the use of LLM will increase the computational load, which is a common challenge faced by the entire community. The results were tested using an A100 GPU, which will likely be lower if a more powerful GPU, such as the H100, is used.

Considering the actual computation time certainly increases latency. However, even with LLM used, the computation-aware latency and translation quality trade-off remain promising. This computation-aware latency result will evolve with hardware advancements, and reducing LLM computational load has been actively studied by the community. 

In addition, comparing the latency results with and without considering the computation time, the smaller $k$ does not make the gap larger than the larger $k$ value.
Smaller k-values will require more frequent past key-value updates before LLM generation after new speech input, as mentioned in Section~\ref{sec:simul-infer}, so it can be seen that this update does not greatly increase the computation time as it is calculated in parallel.
\section{Comparison with Foundation Models}
\label{compare_sota}
\begin{figure}[h]
    \centering
    \includegraphics[width=68mm]{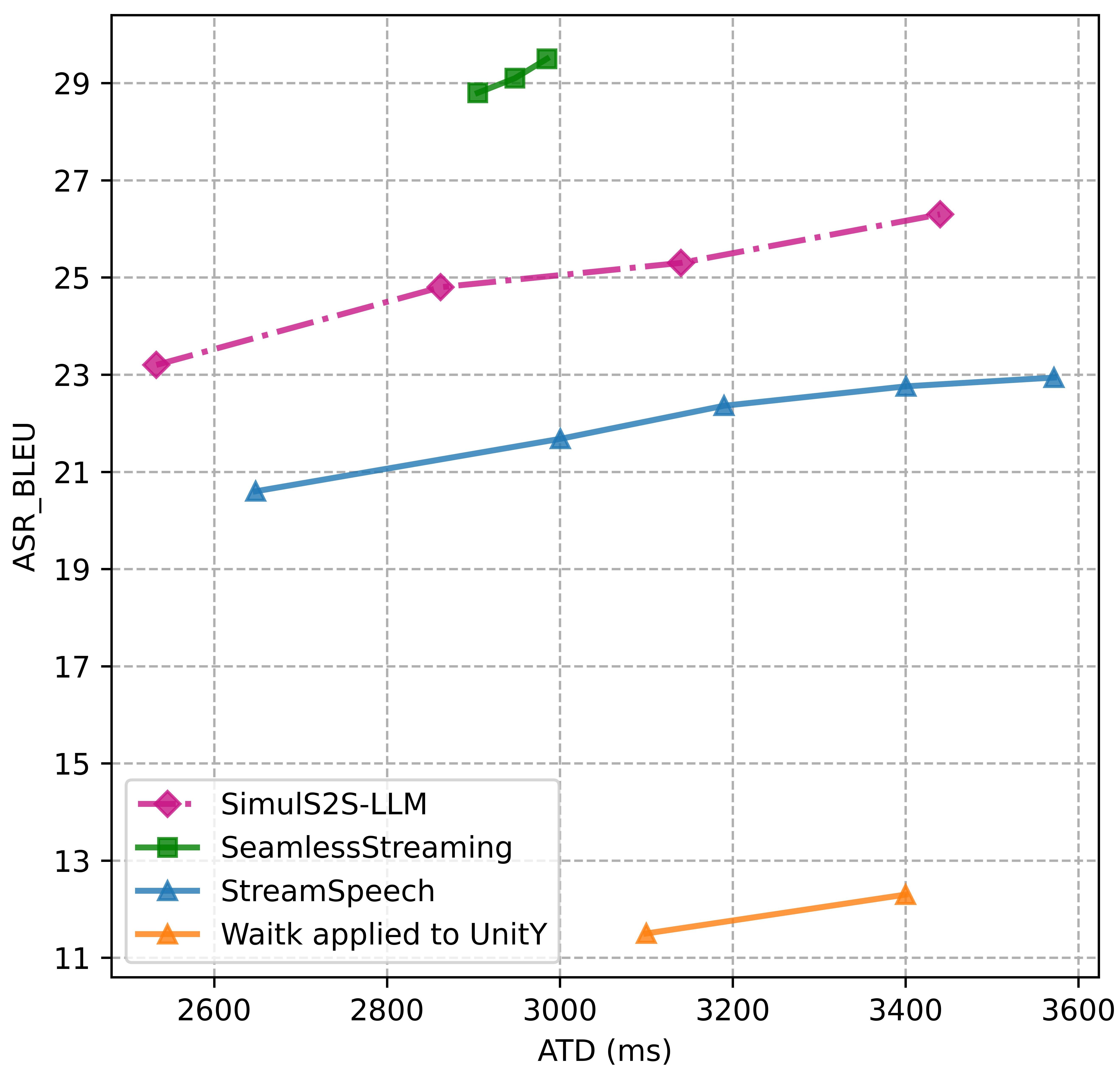}
    \caption{Simul-S2ST results of different models on CVSS-C Es-En. Note the comparison is not well-controlled as the SeamlessStreaming \citep{barrault2023seamless} has been extensively trained as a translation foundation model using about 9,300 hours of speech-to-speech training data. Other models were trained on the same dataset, i.e., CVSS-C, but only SimulS2S-LLM can utilise the LLM.}
    \label{compare-s2st}
\end{figure}

This section further compares the proposed SimulS2S-LLM with the foundation model SeamlessStreaming \citep{barrault2023seamless}, although they are not directly comparable since SeamlessStreaming has been extensively trained on approximately 9,300 hours of speech-to-speech data. Additionally, the results of applying wait-k to the UnitY \cite{DBLP:conf/acl/InagumaPKCWC00023} model, reproduced by \citet{zhang-etal-2024-streamspeech}, are also included for comparison.

\begin{figure}[h]
    \centering
    \includegraphics[width=68mm]{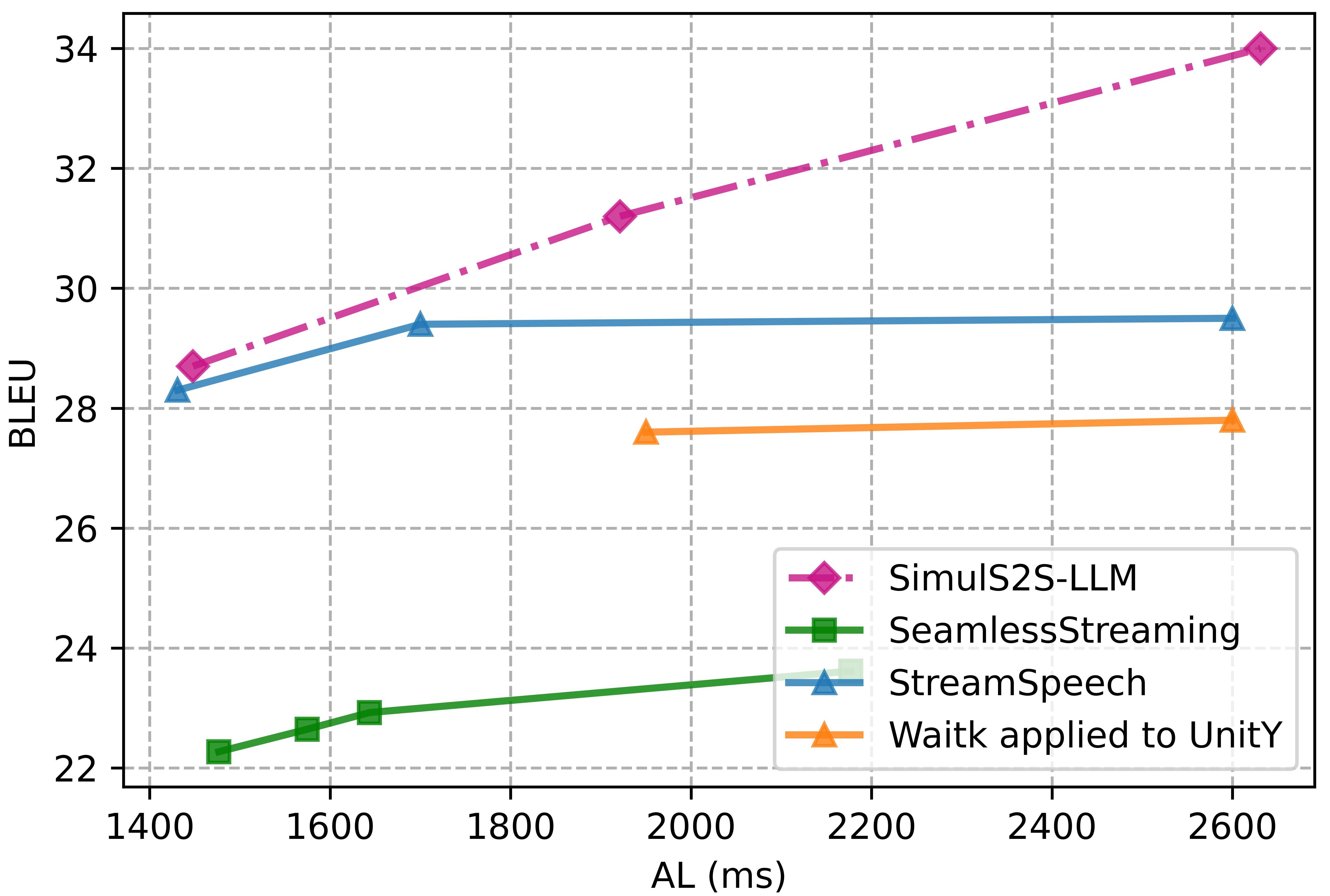}
    \caption{Simul-S2ST results of different models on CVSS-C Es-En. Note the comparison is not well-controlled as Fig.~\ref{compare-s2st}.}
    \label{compare-s2tt}
\end{figure}

As shown in Fig.~\ref{compare-s2st} and Fig.~\ref{compare-s2tt}, StreamSpeech greatly outperforms the UnitY model that uses the Wait-k strategy on both the Simul-S2ST and Simul-S2TT tasks, which is consistent with the findings of \citet{zhang-etal-2024-streamspeech}, showing that StreamSpeech is the existing state-of-the-art solution.

On the Simul-S2ST task, as shown in Fig.~\ref{compare-s2st}, SeamlessStreaming unsurprisingly achieves the best results, but performs poorly on Simul-S2TT as shown in Fig.~\ref{compare-s2tt}, consistent with the findings of \citep{barrault2023seamless}. Therefore, in general, the proposed SimulS2S-LLM achieves promising results, and the gap with SeamlessStreaming on Simul-S2ST is also acceptable considering that SeamlessStreaming uses a much larger training data size. If there is more speech-to-speech training data, the performance of SimulS2S-LLM can be expected to be further greatly improved.

\section{Numerical Values for Figure~\ref{main-s2t}}
\label{numeric-s2t}

The numerical values for Fig.~\ref{main-s2t} are provided in Table~\ref{tab:fisher-laal}. In addition to the AL values displayed in Fig.~\ref{main-s2t}, this table includes the alternative Length-Adaptive Average Lagging (LAAL) \cite{papi-etal-2022-generation} latency metric results. The conclusion remains consistent across different metrics.

\begin{table*}[t]
  \centering
  \setlength{\tabcolsep}{2.5mm}
  \renewcommand\arraystretch{0.97}
  \begin{tabular}{l | c |c c }
    \Xhline{3\arrayrulewidth}
     Simul-S2TT Models&{BLEU}&LAAL(ms)&AL(ms)\\
    \hline
    SimulS2S-LLM (k=3) &{24.7}&1384&{1061}\\
    SimulS2S-LLM (k=4) &{28.7}&1764&{1448}\\
    SimulS2S-LLM (k=5) &{31.2}&2170&{1921}\\
    SimulS2S-LLM (k=7) &{34.0}&2805&{2631}\\
    \hline
    Boundary-unaware SimulS2S-LLM (k=5) &{20.9}&1343&{1042}\\
    Boundary-unaware SimulS2S-LLM (k=7) &{26.4}&1761&{1440}\\
    Boundary-unaware SimulS2S-LLM (k=9) &{27.7}&2317&{2022}\\
    Boundary-unaware SimulS2S-LLM (k=11) &{31.4}&2902&{2704}\\
    \Xhline{3\arrayrulewidth}
  \end{tabular}
  \caption{Numerical Simul-S2TT results of SimulS2S-LLM on CVSS-C Es-En corresponding to Fig.~\ref{main-s2t}.}
  \label{tab:fisher-laal}
\end{table*}
\end{document}